\documentclass[10pt,twocolumn,letterpaper]{article}

\usepackage{iccv}
\usepackage{times}
\usepackage{epsfig}
\usepackage{graphicx}
\usepackage{amsmath}
\usepackage{amssymb}

\usepackage{booktabs}
\usepackage{caption}

\usepackage[breaklinks=true,colorlinks,bookmarks=false]{hyperref}

\iccvfinalcopy 

\def\iccvPaperID{****} 

\ificcvfinal\fi

\begin{document}

\title{Image Inpainting via Conditional Texture and Structure Dual Generation}

\author{Xiefan Guo$^{1,2}$\quad Hongyu Yang$^{2}$\thanks{Corresponding author.}\quad Di Huang$^{1,2}$\\
$^{1}$State Key Laboratory of Software Development Environment, Beihang University, Beijing, China\\
$^{2}$School of Computer Science and Engineering, Beihang University, Beijing, China\\
{\tt\small \{xfguo,hongyuyang,dhuang\}@buaa.edu.cn}
}

\maketitle
\ificcvfinal\thispagestyle{empty}\fi

\begin{abstract}
   Deep generative approaches have recently made considerable progress in image inpainting by introducing structure priors. Due to the lack of proper interaction with image texture during structure reconstruction, however, current solutions are incompetent in handling the cases with large corruptions, and they generally suffer from distorted results. In this paper, we propose a novel two-stream network for image inpainting, which models the \textbf{structure-constrained texture synthesis} and \textbf{texture-guided structure reconstruction} in a coupled manner so that they better leverage each other for more plausible generation. Furthermore, to enhance the global consistency, a {Bi}-directional {G}ated {F}eature {F}usion (Bi-GFF) module is designed to exchange and combine the structure and texture information and a {C}ontextual {F}eature {A}ggregation (CFA) module is developed to refine the generated contents by region affinity learning and multi-scale feature aggregation. Qualitative and quantitative experiments on the CelebA, Paris StreetView and Places2 datasets demonstrate the superiority of the proposed method. Our code is available at \url{https://github.com/Xiefan-Guo/CTSDG}.
\end{abstract}

\vspace{-6pt}

\section{Introduction}
Image inpainting \cite{bertalmio2000image} refers to the process of reconstructing damaged regions of an image while simultaneously maintaining its overall consistency, which is a typical low-level visual task with many practical applications, such as photo editing, distracting object removal, and restoring corrupted parts.

As with most computer vision problems, image inpainting has been largely advanced by the widespread use of deep learning during the past decade. Different from the traditional methods \cite{barnes2009patchmatch, efros2001image} that gradually fill in missing areas by searching for the most similar patches from known regions, the deep generative ones \cite{pathak2016context, iizuka2017globally, yang2017high, yeh2017semantic} capture more high-level semantics and do a better job for images with non-repetitive patterns. There also exists another trend to combine the advantages of deep generative and traditional patch-based methods for image inpainting \cite{yu2018generative, yan2018shift, song2018contextual, liu2019coherent}, delivering inpainting contents with both realistic textures and plausible semantics. Moreover, updated versions of vanilla convolution are investigated \cite{liu2018image, xie2019image, yu2019free}, where operations are masked and normalized to be conditioned only on valid pixels, achieving promising performance for irregular corruptions. Nevertheless, the methods above expose a common drawback in recovering the global structure of the image, as a generative network is not as powerful as expected for this issue.

\begin{figure}
	\centering
	\setlength{\belowcaptionskip}{-0.65cm}
	\setlength{\abovecaptionskip}{0.1cm}
	\includegraphics[width=0.97\linewidth]{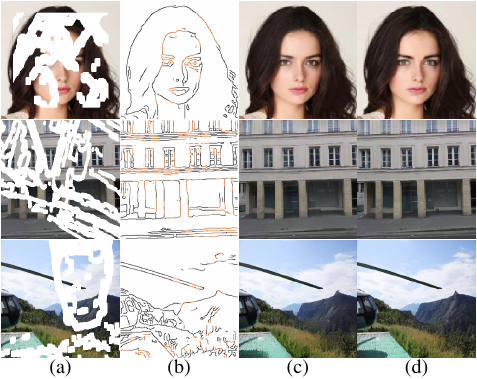}
	\caption{High-quality inpainting results. From left to right: (a) input corrupted images, (b) our reconstructed structures, (c) our filled results, and (d) ground-truth images.} 
	\label{fig:title}
\end{figure}

To deal with this problem, a number of multi-stage methods are proposed to explicitly incorporate structure modeling, which hallucinate structures of missing regions in the first stage and use them to guide pixel generation in the second stage. For instance, EdgeConnect \cite{nazeri2019edgeconnect} encodes such structures by edges, while \cite{ren2019structureflow} and \cite{xiong2019foreground} adopt intermediate edge-preserved smooth images and foreground contours. These alternatives show the results with improved structures and textures. Unfortunately, acquiring reasonable edges from corrupted images is itself a very challenging task, and taking unstable structural priors tends to incur large errors in those series-coupled frameworks.

More recently, a few attempts mix the modeling processes of structures and textures. PRVS (Progressive Reconstruction of Visual Structure) \cite{li2019progressive} and MED (Mutual Encoder-Decoder) \cite{liu2020rethinking} are the representatives, and they generally exploit a shared generator for both textures and structures. Despite some performance gains reported, the relationship between structures and textures is not fully considered in this single entangling architecture. In particular, since image structures and textures correlate throughout the network, it is difficult for them to convey holistic complementary information to assist the other side. Such a fact indicates that there is still much space for improvement.

In this paper, we propose a novel two-stream network which casts image inpainting into two collaborative subtasks, \emph{i.e.}, {structure-constrained texture synthesis} and {texture-guided structure reconstruction}. In this way, the two parallel-coupled streams are individually modeled and combined to complement each other. Correspondingly, a two-branch discriminator is developed to estimate the performance of this generation, which supervises the model to synthesize realistic pixels and sharp edges simultaneously for global optimization. In addition, we introduce a novel Bi-directional Gated Feature Fusion (Bi-GFF) module to integrate the rebuilt structure and texture feature maps to enhance their consistency, along with a Contextual Feature Aggregation (CFA) module to highlight the clues from distant spatial locations to render finer details. Due to the dual generation network as well as the specifically designed modules, our approach is able to achieve more visually convincing structures and textures (see Figure~\ref{fig:title}, zoom in for a better view).

Experiments are extensively conducted on the CelebA \cite{liu2015deep}, Paris StreetView \cite{doersch2012what} and Places2 \cite{zhou2018places} datasets for evaluation. Qualitative and quantitative results demonstrate that our model significantly outperforms the state-of-the-art.

The main novelties and contributions are as follows:

\begin{itemize}
	\item We propose a novel two-stream network for image inpainting, which  models structure-constrained texture synthesis and texture-guided structure reconstruction in a coupled manner so that the dual generation tasks better facilitate each other for more accurate results.
	\item We design a Bi-directional Gated Feature Fusion (Bi-GFF) module to share and combine information between the structure and texture features for consistency enhancement and a Contextual Feature Aggregation (CFA) module to yield more vivid details by modeling long-term spatial dependency.
	\item We achieve the new state-of-the-art performance on multiple public benchmarks both qualitatively and quantitatively.
\end{itemize}

\begin{figure*}
	\centering
	\setlength{\belowcaptionskip}{-0.6cm}
	\setlength{\abovecaptionskip}{0.1cm}
	\includegraphics[width=0.97\textwidth]{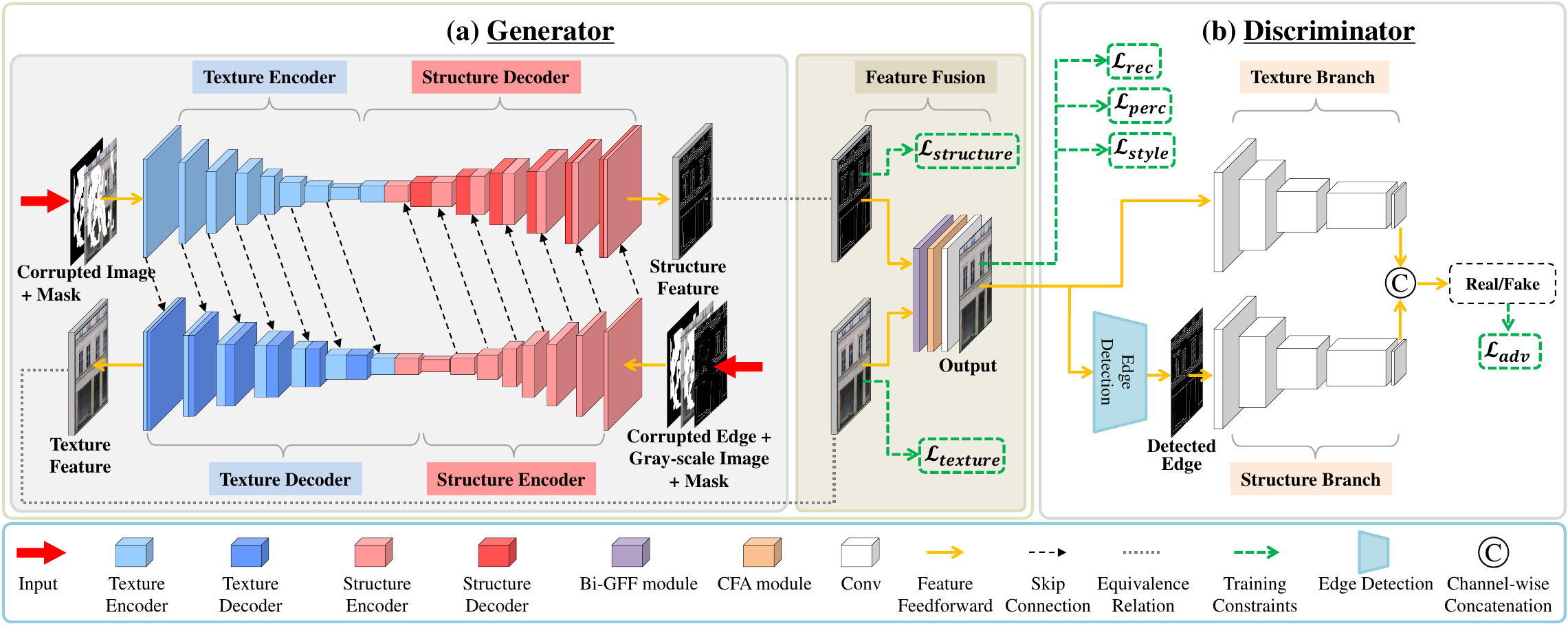}
	\caption{Overview of the proposed method (best viewed in color). \underline{{\bf Generator}}: Image inpainting is cast into two subtasks, \emph{i.e.}, \emph{structure-constrained texture synthesis} (left, {blue}) and \emph{texture-guided structure reconstruction} (right, {red}), and the two parallel-coupled streams borrow encoded deep features from each other. The Bi-directional Gated Feature Fusion (Bi-GFF) module and Contextual Feature Aggregation (CFA) module are stacked at the end of the generator to further refine the results. \underline{{\bf Discriminator}}: The texture branch estimates the generated texture, while the structure branch guides structure reconstruction.}
	\label{fig:generator}
\end{figure*}

\section{Related Work}

\subsection{Traditional Methods}

The traditional methods can be mainly summarized into two categories, \emph{i.e.}, diffusion-based and patch-based. Diffusion-based methods \cite{bertalmio2000image, ballester2001filling} render missing regions referring to the appearance information of the neighboring ones. Their results are not so good due to this preliminary searching mechanism. In patch-based methods \cite{barnes2009patchmatch, xu2010image}, pixel completion is conducted by searching and pasting the most similar patches from undamaged regions of source images, which takes advantage of long-distance information. These methods achieve better performance, but they are computationally expensive when calculating patch similarities between missing and available regions and struggle to reconstruct patterns with rich semantics.

\subsection{Deep Generative Methods}

The deep generative methods \cite{yu2018generative, yu2019free, lahiri2020prior, yi2020contextual, zhao2020uctgan, zhou2020learning, zeng2020high, wang2020vcnet, liao2020guidance} are currently dominating, which effectively extract meaningful semantics from damaged images and recover reasonable contents with high visual fidelity, owing to their powerful feature learning ability.

Recently, Wang \emph{et al.} \cite{wang2018high} significantly improve the quality of image synthesis with sharper edges by involving structural information. Subsequently,
a number of multi-stage methods that serially incorporate additional structural priors are proposed, producing more impressive results. EdgeConnect \cite{nazeri2019edgeconnect} extracts image structures by edges, based on which the holes are filled. Xiong \emph{et al.} \cite{xiong2019foreground} show a similar model while it employs foreground object contours as structure priors instead of edges. Ren \emph{et al.} \cite{ren2019structureflow} point out that edge-preserved smooth images convey better global structure since more semantics are captured. But these methods are sensitive to the accuracy of structures (\emph{e.g.} edges and contours) which is not easy to guarantee. To overcome this weaknesses, several methods attempt to exploit the correlation of textures and structures. Li \emph{et al.} \cite{li2019progressive} design a visual structure reconstruction layer to progressively entangle the generation of image contents and structures. Yang \emph{et al.} \cite{yang2020learning} introduce a multi-task framework to generate sharp edges by adding structural constraints.  Liu \emph{et al.} \cite{liu2020rethinking} present a mutual encoder-decoder network to simultaneously learn the CNN features that correspond to structures and textures with different layers. However, it is rather difficult to model both textures and structures and make them sufficiently complement each other in a single shared architecture.

Our study also makes use of image structural information and figures out a different but more effective two-stream network, where structure-constrained texture synthesis and texture-guided structure reconstruction are jointly considered. The two subtasks better facilitate each other, leading to more convincing textures and structures in dual generation.

\vspace{-4pt}
\section{Approach}

As illustrated in Figure~\ref{fig:generator}, the proposed method is implemented as a generative adversarial network, where the two-stream generator jointly synthesizes image textures and structures, and the discriminator judges their quality and consistency. In this section, we detailedly describe the generator, the discriminator, and the loss functions.

\vspace{-3.pt}
\subsection{Generator}

The generator is a two-stream architecture, modeled by a U-Net variant, as shown in Figure~\ref{fig:generator}~(a). At the encoding stage, the corrupted image and its corresponding edge map are individually projected into the latent space, where the left branch focuses on texture features and the right branch targets structure features. At the decoding stage, the texture decoder synthesizes structure-constrained textures by borrowing structure features from the structure encoder, while the structure decoder recovers texture-guided structures by taking texture features from the texture encoder. With such a dual generation architecture, structures and textures well complement each other, leading to improved results. 

In this encoder-decoder based backbone, we replace all the vanilla convolutions with the partial convolution layers to better capture information from irregular boundaries, since partial convolutions are conditioned only on uncorrupted pixels. Besides, skip connections are utilized to produce more sophisticated predictions by combining low-level and high-level features at multiple scales. To enhance the consistency of the rebuilt structures and textures, the feature maps output by the two branches are further fused to render the final result through a specially designed Bi-GFF module followed by a CFA module. Refer to the supplementary material for more details of the backbone.

\begin{figure}
	\centering
	\setlength{\belowcaptionskip}{-0.4cm}
	\setlength{\abovecaptionskip}{0.1cm}
	\includegraphics[width=1.0\linewidth]{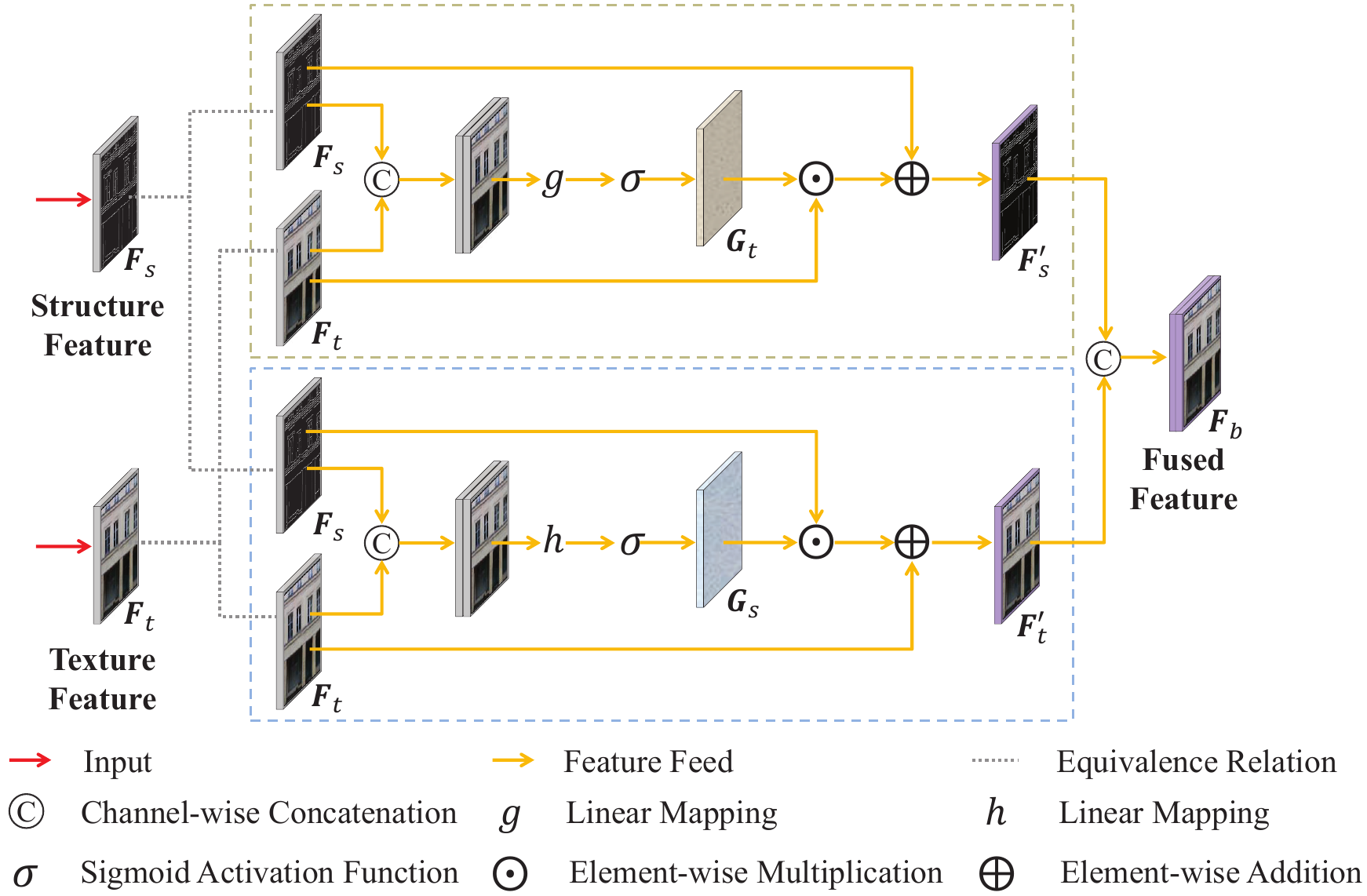}
	\caption{Illustration of the Bi-directional Gated Feature Fusion (Bi-GFF) module, which entangles the decoded structure and texture features to refine the results.}
	\label{fig:sal}
\end{figure}

\noindent \textbf{Bi-directional Gated Feature Fusion (Bi-GFF).} This module is proposed to further combine the decoded texture and structure features. It exchanges messages between the two kinds of information, where soft gating is exploited to control the rate. Due to this integration operation, the feature is refined and simultaneously texture- and structure-aware. Figure~\ref{fig:sal} illustrates the Bi-GFF module. 

Specifically, the texture feature map output by the decoder is denoted as $\boldsymbol{F}_{t}$ and the structure feature map is denoted as $\boldsymbol{F}_{s}$. To build texture-aware structure features, a soft gating $\boldsymbol{G}_{t}$, which controls to what extent the texture information is integrated, is formulated as:
\begin{equation}
\setlength\abovedisplayskip{5pt}
\setlength\belowdisplayskip{5pt}
\boldsymbol{G}_{t} = {\sigma}\left(g\left ( \text{Concat}\left (\boldsymbol{F}_{t}, \boldsymbol{F}_{s} \right)\right)\right),
\end{equation}
where $\text{Concat}(\cdot)$ is channel-wise concatenation, $g(\cdot)$ is the mapping function implemented by a convolution layer with the kernel size of 3, and $\sigma(\cdot)$ is Sigmoid activation. With $\boldsymbol{G}_{t}$, we adaptively merge $\boldsymbol{F}_{t}$ into $\boldsymbol{F}_{s}$ as:
\begin{equation}
\setlength\abovedisplayskip{4pt}
\setlength\belowdisplayskip{4pt}
\boldsymbol{F}_{s}^{'} = \alpha(\boldsymbol{G}_{t} \odot \boldsymbol{F}_{t})\oplus \boldsymbol{F}_{s},
\end{equation}
where $\alpha$ is a training parameter initialized to zero, and $\odot$ and $\oplus$ denote element-wise multiplication and element-wise addition, respectively.

Symmetrically, we calculate the structure-aware texture feature $\boldsymbol{F}_{t}^{'}$ as follows:
\begin{align}
\boldsymbol{G}_{s} &= {\sigma}\left(h\left ( \text{Concat}\left (\boldsymbol{F}_{t}, \boldsymbol{F}_{s} \right)\right)\right), \\
\boldsymbol{F}_{t}^{'} &= \beta(\boldsymbol{G}_{s} \odot \boldsymbol{F}_{s})\oplus \boldsymbol{F}_{t},
\end{align}
where $h$ follows the same pattern as $g$ and $\beta$ is a training parameter initialized to zero as $\alpha$.

Finally, we fuse $\boldsymbol{F}_{s}^{'}$ and $\boldsymbol{F}_{t}^{'}$ to obtain the integrated feature map $\boldsymbol{F}_{b}$ by channel-wise concatenation:
\begin{equation}
{\boldsymbol{F}}_{b} = \text{Concat}(\boldsymbol{F}_{s}^{'}, \boldsymbol{F}_{t}^{'}).
\end{equation}

\begin{figure}
	\centering
	\setlength{\belowcaptionskip}{-0.4cm}
	\setlength{\abovecaptionskip}{0.1cm}
	\includegraphics[width=1.0\linewidth]{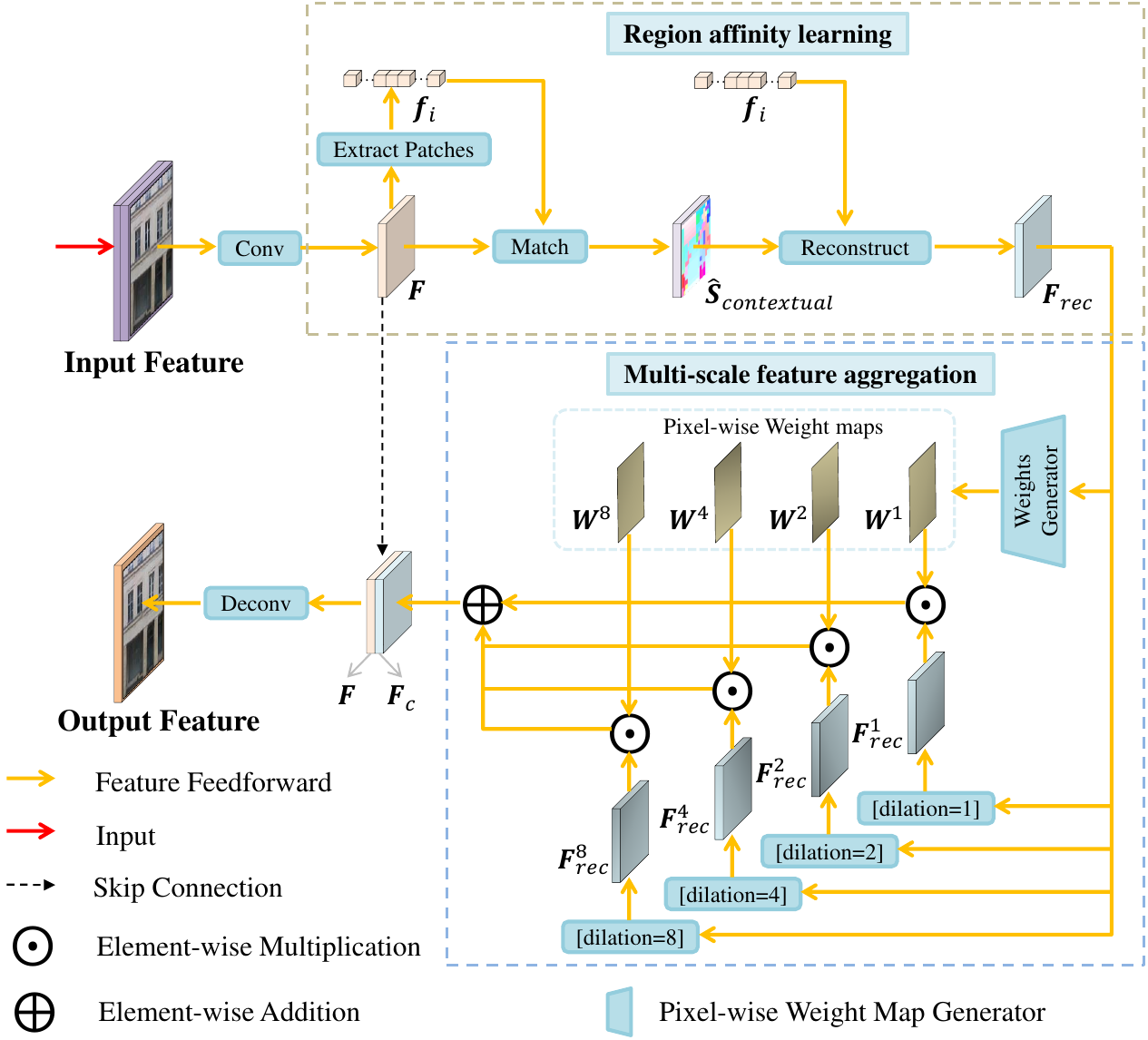}
	\caption{Illustration of the Contextual Feature Aggregation (CFA) module, which models long-term spatial dependency by capturing features at diverse semantic levels.}
	\label{fig:cal}
\end{figure}

\noindent \textbf{Contextual Feature Aggregation (CFA).} To better learn which existing regions contribute to filling holes, this module is designed, which enhances the correlation between local features of an image and maintains the overall image consistency. It is inspired by \cite{yu2018generative}, but unlike its fixed-scale patch matching scheme, in this study, multi-scale feature aggregation is adopted to encode rich semantic features at multiple scales so that it well balances the accuracy and complexity to handle more challenging cases, in particular, scale changes. The detailed process is depicted in Figure~\ref{fig:cal}.

To be specific, given a feature map $\boldsymbol{F}$, we first extract the patches of $3\times 3$ pixels and calculate their cosine similarities as:
\begin{equation}
\boldsymbol{S}_{contextual}^{i,j} = \left \langle \frac{\boldsymbol{f}_i}{{\| \boldsymbol{f}_i \|}_{2}}, \frac{\boldsymbol{f}_j}{{\| \boldsymbol{f}_j \|}_{2}} \right \rangle,
\end{equation}
where $\boldsymbol{f}_i$ and $\boldsymbol{f}_j$ correspond to the $i$-th and $j$-th patch of the feature map, respectively.

We then apply softmax to the similarities to obtain the attention score of each patch:
\begin{equation}
\hat{\boldsymbol{S}}_{contextual}^{i,j} = \frac{\exp\left(\boldsymbol{S}_{contextual}^{i,j}\right)}{\sum_{j=1}^{N} \exp\left(\boldsymbol{S}_{contextual}^{i,j}\right)}.
\end{equation}

Next, the extracted patches are reused to reconstruct the feature map based on the attention map:
\begin{equation}
\tilde{\boldsymbol{f}}_{i} = \sum_{j=1}^{N} \boldsymbol{f}_{j} \cdot \hat{\boldsymbol{S}}_{contextual}^{i,j},
\end{equation}
where $\tilde{\boldsymbol{f}}_{i}$ is the $i$-th patch of the reconstructed feature map $\boldsymbol{F}_{rec}$.  The operations above are implemented as convolution, channel-wise softmax, and deconvolution, respectively.
  
When the feature map is reconstructed, four sets of dilated convolution layers with different dilation rates are used to capture multi-scale semantic features:
\begin{equation}
{\boldsymbol{F}}_{rec}^{k} = \text{Conv}_{k} \left( \boldsymbol{F}_{rec} \right),
\end{equation}
where $\text{Conv}_{k}(\cdot)$ denotes dilated convolution layers with dilation rate of $k$, $k \in \{1, 2, 4, 8\}$.

To better aggregate the multi-scale semantic features, we further design a pixel-level weight map generator $G_{w}$, which aims to predict the pixel-wise weight maps. In our implementation, $G_{w}$ consists of two convolution layers with the kernel size of 3 and 1, respectively, each of which is followed by ReLU non-linear activation, and the number of the output channels for $G_{w}$ is set to 4. The pixel-wise weight maps are calculated as:
\begin{align}
\boldsymbol{W} = \text{Softmax} \left( G_{w} \left( \boldsymbol{F}_{rec} \right) \right), \\
\boldsymbol{W}^1, \boldsymbol{W}^2, \boldsymbol{W}^4, \boldsymbol{W}^8 = \text{Slice}(\boldsymbol{W}),
\end{align}
where $\text{Softmax}(\cdot)$ is channel-wise softmax and $\text{Slice}(\cdot)$ is channel-wise slice. Finally, the multi-scale semantic features are aggregated to produce the refined feature map $\boldsymbol{F}_{c}$ by element-wise weighted sum:
\begin{equation}
\begin{aligned}
\boldsymbol{F}_{c} =& \left( \boldsymbol{F}_{rec}^{1} \odot \boldsymbol{W}^1 \right) \oplus \left( \boldsymbol{F}_{rec}^{2} \odot \boldsymbol{W}^2 \right) \oplus \\
& \left( \boldsymbol{F}_{rec}^{4} \odot \boldsymbol{W}^4 \right) \oplus \left( \boldsymbol{F}_{rec}^{8} \odot \boldsymbol{W}^8 \right).
\end{aligned}
\end{equation}

Note, as the mask update mechanism of partial convolution layers is exploited, there is no need to distinguish the foreground and background pixels of the image as \cite{yu2018generative} does. Skip connection \cite{ronneberger2015u} is adopted to prevent semantic damage caused by patch-shift operations and a pair of convolution and deconvolution layers are seamlessly embedded into our architecture to improve computational efficiency.

\subsection{Discriminator}
 
Motivated by global and local GANs \cite{iizuka2017globally}, Gated Convolution \cite{yu2019free} and Markovian GANs \cite{li2016precomputed}, we develop a two-stream discriminator to distinguish genuine images from the generated ones by estimating the feature statistics of both texture and structure. The discriminator is shown in Figure~\ref{fig:generator}~(b). The texture branch includes three convolution layers with the kernel size of 4 and stride of 2, tailed by two convolution layers with the kernel size of 4 and stride of 1. We use the Sigmoid non-linear activation function at the last layer and the Leaky ReLU with the slope of 0.2 for other layers. The structure branch shares the same pattern as the upper stream, where the input edge map is detected by a residual block \cite{he2016deep} followed by a convolution layer with the kernel size of 1. Finally, the outputs of the two branches are concatenated in the channel dimension, based on which we calculate the adversarial loss.

Different from the case in the texture branch, it is intractable to optimize the adversarial loss of the structure branch only with the detected edge map, mainly due to the sparse nature of the edge. We therefore adopt the gray-scale image as an additional condition and feed the paired data as the input in the structure branch, as several previous studies do \cite{xiong2019foreground, nazeri2019edgeconnect}. As such, the structure branch not only estimates the authenticity of the generated structure, but also guarantees its consistency with the ground-truth image. Besides, spectral normalization \cite{miyato2018spectral} is used, as it proves effective in solving the well-known training instability problem of generative adversarial networks.

\subsection{Loss Functions}

The model is trained with a joint loss, containing the reconstruction loss, perceptual loss, style loss and adversarial loss, to render visually realistic and semantically reasonable results.

Formally, let $G$ be the generator and $D$ be the discriminator. Denote by $\boldsymbol{I}_{gt}$ the ground-truth image, $\boldsymbol{E}_{gt}$ the complete edge map, $\boldsymbol{Y}_{gt}$ the gray-scale image, $\boldsymbol{M}_{in}$ the initial binary mask (with value 1 for existing region, 0 otherwise), $\boldsymbol{I}_{in} = \boldsymbol{I}_{gt} \odot \boldsymbol{M}_{in}$ the damaged image, $\boldsymbol{E}_{in} = \boldsymbol{E}_{gt} \odot \boldsymbol{M}_{in}$ the damaged edge map, and $\boldsymbol{Y}_{in} = \boldsymbol{Y}_{gt} \odot \boldsymbol{M}_{in}$ the damaged gray-scale image. The output of our generator is defined as $\boldsymbol{I}_{out}, \boldsymbol{E}_{out} = G(\boldsymbol{I}_{in}, \boldsymbol{E}_{in}, \boldsymbol{Y}_{in}, \boldsymbol{M}_{in})$.

\noindent \textbf{Reconstruction Loss.} We adopt the ${\ell}_1$ distance between $\boldsymbol{I}_{out}$ and $\boldsymbol{I}_{gt}$ as the reconstruction loss, formulated as:
\begin{equation}
\mathcal{L}_{rec} = \mathbb{E} \left [ {\left\| \boldsymbol{I}_{out} - \boldsymbol{I}_{gt} \right\|}_1 \right ].
\end{equation}

\noindent \textbf{Perceptual Loss.} Since the reconstruction loss struggles to capture high-level semantics, we introduce the perceptual loss $\mathcal{L}_{perc}$ to evaluate the global
structure of an image. It measures the ${\ell}_1$ distance of $\boldsymbol{I}_{out}$ to $\boldsymbol{I}_{gt}$ in the feature space defined by the VGG-16 network \cite{simonyan2015very} pre-trained on ImageNet \cite{russakovsky2015imagenet}:
\begin{equation}
\mathcal{L}_{perc} = \mathbb{E} \left[ \sum_{i} {\left\| \phi_i \left( \boldsymbol{I}_{out} \right) - \phi_i \left( \boldsymbol{I}_{gt} \right) \right\|}_1 \right],
\end{equation}
where  ${\phi}_i(\cdot)$ denotes the activation map of the $i$-th pooling layer from VGG-16 given the input image $\boldsymbol{I}_{*}$. In our implementation, \emph{pool}-1, \emph{pool}-2 and \emph{pool}-3 are used.

\begin{figure*}
	\centering
	\setlength{\belowcaptionskip}{-0.4cm}
	\setlength{\abovecaptionskip}{0.1cm}
	\includegraphics[width=1\linewidth]{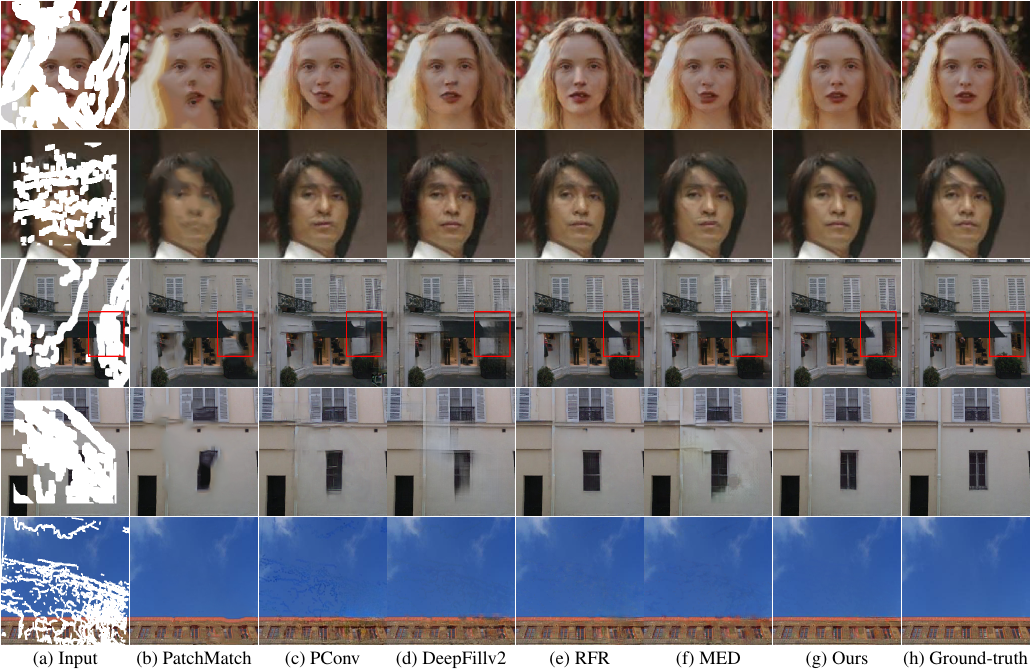}
	\caption{Qualitative comparison on CelebA, Paris StreetView and Places2 (zoom in for a better view): (a) input corrupted images, (b) PatchMatch \cite{barnes2009patchmatch}, (c) PConv \cite{liu2018image}, (d) DeepFillv2 \cite{yu2019free}, (e) RFR \cite{li2020recurrent}, (f) MED \cite{liu2020rethinking}, (g) Ours, and (h) ground-truth images.} 
	\label{fig:compare}
\end{figure*}

\noindent \textbf{Style Loss.} We further include the style loss to ensure style consistency. Similarly, the style loss calculates the ${\ell}_1$ distance between feature maps:
\begin{equation}
\mathcal{L}_{style} = \mathbb{E} \left[ \sum_{i} {\left\| \left(\psi_i \left( \boldsymbol{I}_{out} \right) - \psi_i \left( \boldsymbol{I}_{gt} \right) \right) \right\|}_1 \right],
\end{equation}
where ${\psi}_i(\cdot) = {\phi_i(\cdot)}^T{\phi_i(\cdot)}$ denotes the Gram matrix constructed from the activation map $\phi_i$.

\noindent \textbf{Adversarial Loss.} The adversarial loss is to guarantee the visual authenticity of the reconstructed image as well as the consistency of textures and structures, defined as:
\begin{equation}
\begin{aligned}
\mathcal{L}_{adv} = \min_G &\max_D {\mathbb{E}}_{\boldsymbol{I}_{gt}, \boldsymbol{E}_{gt}} \left[ \log D\left( \boldsymbol{I}_{gt}, \boldsymbol{E}_{gt} \right) \right]\\
&+ {\mathbb{E}}_{\boldsymbol{I}_{out}, \boldsymbol{E}_{out}} \log\left[ 1 - D\left( \boldsymbol{I}_{out}, \boldsymbol{E}_{out} \right) \right].
\end{aligned}
\end{equation}

\noindent \textbf{Intermediate Loss.} To encourage the structure and texture features to be accurately captured by the two decoders, respectively, we introduce intermediate supervisions on $\boldsymbol{F}_s$ and $\boldsymbol{F}_t$:\begin{equation}
\begin{aligned}
\mathcal{L}_{inter} &= \mathcal{L}_{structure} + \mathcal{L}_{texture} \\
&= \text{BCE}(\boldsymbol{E}_{gt}, \mathcal{P}_s(\boldsymbol{F}_s)) + \ell_1(\boldsymbol{I}_{gt}, \mathcal{P}_t(\boldsymbol{F}_t)),
\end{aligned}
\end{equation}
where $\mathcal{P}_s$ and $\mathcal{P}_t$ denote the projection functions implemented by a residual block followed by a convolution layer, which map $\boldsymbol{F}_s$ and $\boldsymbol{F}_t$ to edge map and RGB image, respectively.

In summary, the joint loss is written as:
\begin{equation}
\begin{aligned}
\mathcal{L}_{joint} &= \lambda_{rec}\mathcal{L}_{rec} + \lambda_{perc} \mathcal{L}_{perc} + \lambda_{style} \mathcal{L}_{style} \\
&+ \lambda_{adv}\mathcal{L}_{adv} + \lambda_{inter} \mathcal{L}_{inter},
\end{aligned}
\end{equation}
where $\lambda_{rec}$, $\lambda_{perc}$, $\lambda_{style}$, $\lambda_{adv}$ and $\lambda_{inter}$ are the tradeoff parameters, and we empirically set $\lambda_{rec}=10$, $\lambda_{perc}=0.1$, $\lambda_{style}=250$, $\lambda_{adv}=0.1$, and $\lambda_{inter}=1$.

\section{Experiments}

Extensive experiments are conducted on three public datasets for both subjective and objective evaluation. Ablation studies are also performed to validate the specifically designed architecture and modules.

\subsection{Experimental Settings}

We evaluate the proposed method on the CelebA \cite{liu2015deep}, Paris StreetView \cite{doersch2012what} and Places2 \cite{zhou2018places} datasets, which are widely adopted in the literature, and we follow their original training, testing, and validation splits. Irregular masks are obtained from \cite{liu2018image} and classified based on their hole sizes relative to the entire image with an increment of 10\%. All the images and corresponding masks are resized to $256\times256$ pixels.

\begin{figure*}
	\centering
	\setlength{\belowcaptionskip}{-0.2cm}
	\setlength{\abovecaptionskip}{0.1cm}
	\includegraphics[width=1\linewidth]{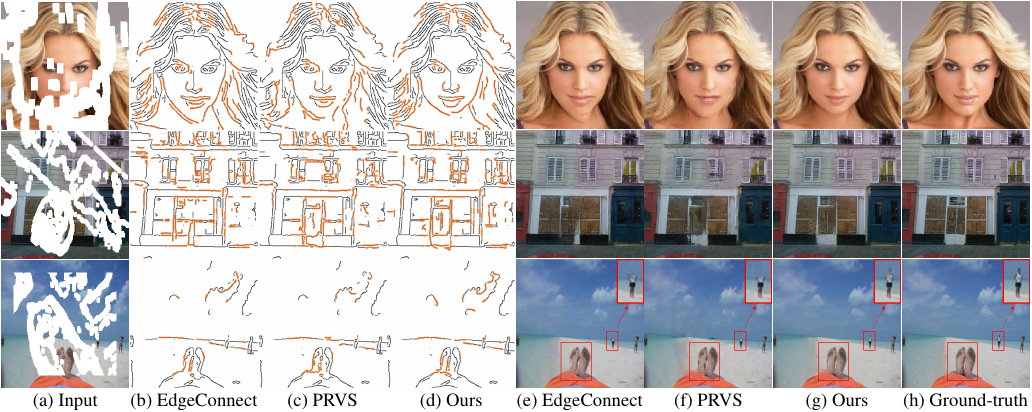}
	\caption{Visual comparison of different structure-based methods on CelebA, Paris StreetView and Places2 (zoom in for a better view): (a) input corrupted images; (b, c, d) reconstructed structures of EdgeConnect \cite{nazeri2019edgeconnect}, PRVS \cite{li2019progressive} and Ours; (e, f, g) corresponding filled results of EdgeConnect \cite{nazeri2019edgeconnect}, PRVS \cite{li2019progressive} and Ours; and (h) ground-truth images.} 
	\label{fig:edge}
\end{figure*}

\begin{table*}[!t]
	\centering
	\setlength{\belowcaptionskip}{-0.5cm} 
	\setlength{\abovecaptionskip}{0.1cm}
	\scalebox{0.9}{
	\begin{tabular}{l|ccc|ccc|ccc|c}
		\toprule
		\multicolumn{1}{c|}{Metrics} & \multicolumn{3}{c|}{LPIPS$^{\dag}$} & \multicolumn{3}{c|}{PSNR$^{\P}$} & \multicolumn{3}{c|}{SSIM$^{\P}$} & User Study$^{\P}$ \\
		\midrule
		\multicolumn{1}{c|}{Mask Ratio} & {\normalsize 0-20\%} & {\normalsize 20-40\%} & {\normalsize 40-60\%} & {\normalsize 0-20\%} & {\normalsize 20-40\%} & {\normalsize 40-60\%} & {\normalsize 0-20\%} & {\normalsize 20-40\%} & {\normalsize 40-60\%} & {\normalsize 0-60\%} \\
		\midrule
		\midrule
		PatchMatch \cite{barnes2009patchmatch} & 0.074 & 0.183 & 0.332 & 30.02 & 24.77 & 20.51 & 0.864 & 0.680 & 0.487 & 2.7\% \\
		PConv \cite{liu2018image} & 0.065 & 0.134 & 0.283 & 30.19 & 25.18 & 21.20 & 0.885 & 0.730 & 0.527 & 4.0\% \\
		DeepFillv2 \cite{yu2019free} & 0.056 & 0.123 & 0.266 & 30.32 & 25.34 & 21.48 & 0.889 & 0.735 & 0.531 & 14.0\% \\
		RFR \cite{li2020recurrent} & 0.048 & 0.101 & 0.239 & 30.74 & 25.80 & 21.99 & 0.899 & 0.750 & 0.553 & 23.3\% \\
		\midrule
		EdgeConnect \cite{nazeri2019edgeconnect} & 0.061 & 0.131 & 0.268 & 30.28 & 25.30 & 21.39 & 0.886 & 0.737 & 0.535 & 4.7\% \\
		PRVS \cite{li2019progressive} & 0.057 & 0.124 & 0.257 & 30.30 & 25.39 & 21.50 & 0.893 & 0.742 & 0.541 & 5.3\% \\
		MED \cite{liu2020rethinking} & 0.053 & 0.120 & 0.248 & 30.41 & 25.45 & 21.63 & 0.895 & 0.745 & 0.547 & 6.0\% \\
		Ours & {\bf 0.042} & {\bf 0.095} & {\bf 0.227} & {\bf 30.81} & {\bf 25.97} & {\bf 22.23} & {\bf 0.904} & \textbf{0.759} & {\bf 0.561} & {\bf 40.0\%} \\
		\bottomrule
	\end{tabular}
	}
	\caption{Objective quantitative comparison and user study on Places2 ($^{\dag}$Lower is better; $^{\P}$Higher is better).}
	\label{tab:quantitative-comparison}
\end{table*}

The model is implemented in PyTorch. Training is launched on a single NVIDIA 1080TI GPU (11GB) with the batch size of 6, optimized with the Adam optimizer. Analogous to \cite{liu2018image}, we first use a learning rate of $2\times 10^{-4}$ for initial training, then finetune the model with a learning rate of $5 \times 10^{-5}$, and freeze the Batch Normalization (BN) parameters of the generator. The discriminator is trained with a learning rate of 1/10 of the generator. It takes around 4 days to train the models on CelebA and Paris StreetView and 10 days on Places2. The fine-tuning is completed within one day. The detailed architectures of the networks are shown in the supplementary material.
 
\subsection{Qualitative Comparison}

\begin{figure*}
	\centering
	\setlength{\belowcaptionskip}{-0.1cm}
	\setlength{\abovecaptionskip}{0.1cm}
	\includegraphics[width=1\linewidth]{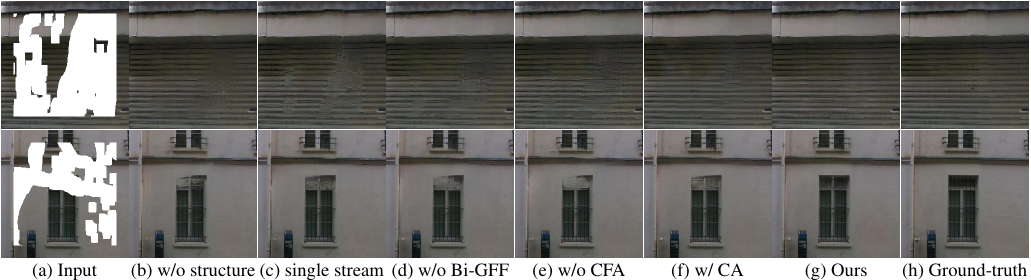}
	\caption{Visualization of the effects of network architecture and individual modules on Paris StreetView.}
	\label{fig:ablation}
\end{figure*}

Figure~\ref{fig:compare} compares our results with the ones of the representative methods including the current state-of-the-arts on the three benchmarks. It can be seen, as a classical patch-based method, PatchMatch \cite{barnes2009patchmatch} fails in handling large holes. PConv \cite{liu2018image} is suitable for irregular corruptions, but obvious artifacts can be observed in Figure~\ref{fig:compare}~(c). DeepFilllv2 \cite{yu2019free} suffers from over-smoothing predictions and distorted structures. With the Recurrent Feature Reasoning module, RFR \cite{li2020recurrent} yields competitive results; however, the details are still not so elegant as ours (the face and sky in Figure~\ref{fig:compare} serve as examples). MED \cite{liu2020rethinking} attempts to correlate structure and texture generation, while the shared generator is inadequate for generating sharp edges and clear textures (\emph{e.g.}, facades in Figure~\ref{fig:compare}).

We also show additional comparison to EdgeConnect \cite{nazeri2019edgeconnect} and PRVS \cite{li2019progressive} in Figure~\ref{fig:edge} as these methods all claim to improve results by reconstructing image structures. Comparatively, the proposed model recovers more reasonable and sharper structures, leading to better results.

To sum up, our model is able to hallucinate more meaningful structures and vivid textures through dual generation.

\subsection{Quantitative Comparison}

\noindent \textbf{Objective evaluation.} We quantitatively evaluate the proposed method using three major metrics: LPIPS, PSNR and SSIM, and compare the scores to those of the state-of-the-art counterparts with irregular mask ratios of 0-20\%, 20-40\% and 40-60\%. Table~\ref{tab:quantitative-comparison} shows the results achieved on the Places2 dataset, where the proposed method outperforms the other approaches, clearly demonstrating its effectiveness. More comparison on the CelebA and Paris StreetView datasets is shown in the supplementary material.

\begin{table*}[!t]
	\centering
	\setlength{\belowcaptionskip}{-0.45cm}
	\setlength{\abovecaptionskip}{0.1cm}
	\scalebox{0.95}{
		\begin{tabular}{c|ccc|ccc|ccc}
			\toprule
			\multicolumn{1}{c|}{Metrics} & \multicolumn{3}{c|}{LPIPS$^{\dag}$} & \multicolumn{3}{c|}{PSNR$^{\P}$} & \multicolumn{3}{c}{SSIM$^{\P}$} \\
			\midrule
			\multicolumn{1}{c|}{Mask Ratio} & {\normalsize 0-20\%} & {\normalsize 20-40\%} & {\normalsize 40-60\%} & {\normalsize 0-20\%} & {\normalsize 20-40\%} & {\normalsize 40-60\%} & {\normalsize 0-20\%} & {\normalsize 20-40\%} & {\normalsize 40-60\%} \\
			\midrule
			\midrule
			w/o structure priors & 0.054 & 0.129 & 0.251 & 31.72 & 26.71 & 22.22 & 0.909 & 0.755 & 0.550 \\
			single-stream & 0.051 & 0.122 & 0.245 & 32.27 & 27.03 & 22.59 & 0.913 & 0.764 & 0.558 \\
			\midrule
			w/o Bi-GFF & 0.045 & 0.114 & 0.236 & 32.61 & 27.20 & 22.75 & 0.919 & 0.772 & 0.567 \\
			w/o CFA & 0.049 & 0.119 & 0.243 & 32.34 & 27.09 & 22.64 & 0.914 & 0.766 & 0.561 \\
			w/ CA & 0.043 & 0.115 & 0.240 & 32.54 & 27.15 & 22.69 & 0.920 & 0.769 & 0.566 \\
			\midrule
			Ours & {\bf 0.039} & {\bf 0.107} & {\bf 0.226} & {\bf 32.93} & {\bf 27.48} & {\bf 22.89} & {\bf 0.923} & {\bf 0.777} & {\bf 0.573} \\
			\bottomrule
		\end{tabular}
	}
	\caption{Quantitative ablation study on Paris StreetView.}
	\label{tab:ablation}
\end{table*}

\noindent \textbf{User Study.} We further perform subjective user study. 10 volunteers with image processing expertise are involved in this evaluation. They are invited to choose the most realistic image from those inpainted by the proposed method and the representative state-of-the-art approaches. Specifically, each participant has 15 questions, which are randomly sampled from the Places2 dataset. We tally the votes and show the statistics in Table~\ref{tab:quantitative-comparison}. Our method performs more favorably against the other ones by a large margin, clearly validating its effectiveness.

\subsection{Analysis on Network Architecture}

Our model assumptions include that structure priors are essential to image inpainting and dual generation of textures and structures is beneficial. We therefore experimentally verify the credit of such architecture design on Paris StreetView. 

\noindent \textbf{On Structure Priors.} To highlight the structure priors, we build a single-stream network as baseline, which fills missing regions by solely modeling texture features, and the discriminator is single-stream accordingly. As shown in Figure~\ref{fig:ablation}~(b), the baseline method does not well deal with complex structures and tends to smooth out the detailed textures in case of large corruptions. The quantitative results in Table~\ref{tab:ablation} also indicate that our network with structure priors significantly improves the baseline.

\noindent \textbf{On Two-stream Network Architecture.} To further highlight the two-stream dual generation architecture, we compare it with a multi-task single-stream network, which is tailed by two branches to model the image structure and texture simultaneously. We enlarge its channels to make it have the same amount of parameters as the proposed network. The Bi-GFF and CFA modules are embedded to refine generation as the proposed model. As shown in Figure~\ref{fig:ablation}~(c), the two-stream architecture exhibits superior performance with more visually reasonable structures and detailed textures. Quantitative results in Table~\ref{tab:ablation} also validate the advantages of texture and structure dual generation.

\subsection{Ablation Study}

\noindent \textbf{On Bi-directional Gated Feature Fusion.} The Bi-GFF module is developed to enhance the consistency of the rebuilt structures and textures. For the results obtained using a simpler fusion module (a channel-wise concatenation followed by a convolution layer), blurred edges and unexpected noise can be observed in Figure~\ref{fig:ablation}~(d), especially around complex boundaries, such as the windows. To make the comparison more specific, quantitative results are given in Table~\ref{tab:ablation}, which indicate that Bi-GFF contributes to the performance gain. 

\noindent \textbf{On Contextual Feature Aggregation.} The CFA module is introduced to enhance the correlation between local features and the overall image consistency. As shown in Figure~\ref{fig:ablation}~(e), the model without CFA renders low-quality images, and texture filling is sensitive to structure noise. Quantitative results in Table~\ref{tab:ablation} also validate its necessity.

\noindent \textbf{On Multi-scale Feature Aggregation in CFA.} As our CFA module is updated from the contextual attention layer \cite{yu2018generative}, we directly compare it with the original version to prove its effectiveness. As shown in Figure~\ref{fig:ablation}~(f) and Table~\ref{tab:ablation}, we demonstrate that multi-scale feature aggregation obviously benefits the quality of the results, with consistent textures and better quantitative scores reported.

\section{Conclusion}

In this paper, we propose a novel two-stream image inpainting method, which recovers corrupted image by simultaneously modeling structure-constrained texture synthesis and texture-guided structure reconstruction. In this way, the two subtasks exchange useful information and thus facilitate each other. Furthermore, a Bi-directional Gated Feature Fusion module is introduced followed by a Contextual Feature Aggregation module to refine the results, with both semantically reasonable structures and detail-rich textures. Experiments show that this model is competent for this issue and outperforms the state-of-the-art counterparts.

\section*{Acknowledgment} 
This work is partly supported by the National Natural Science Foundation of China (No. 62022011), the China Post-doctoral Science Foundation (No. 2019M660406), the Research Program of State Key Laboratory of Software Development Environment (SKLSDE-2021ZX-04), and the Fundamental Research Funds for the Central Universities. We also give specical thanks to Alibaba Group for their contribution to this paper.

{\small
\bibliographystyle{ieee_fullname}
\bibliography{egbib}
}

\end{document}